\documentclass[journal]{IEEEtran}
\usepackage{xcolor,soul,framed} 
\colorlet{shadecolor}{yellow}
\usepackage[pdftex]{graphicx}
\graphicspath{{../pdf/}{../jpeg/}}
\DeclareGraphicsExtensions{.pdf,.jpeg,.png}
\usepackage[cmex10]{amsmath}
\usepackage{array}
\usepackage{mdwmath}
\usepackage{mdwtab}
\usepackage{eqparbox}
\usepackage{mathtools}
\usepackage{url}
\usepackage{amsfonts}
\usepackage{longtable}
\usepackage{pdflscape}
\usepackage{rotating} 
\usepackage{lipsum} 

\begin{document}
\bstctlcite{IEEEexample:BSTcontrol}
    \title{Efficient Learning Content Retrieval with Knowledge Injection}
  \author{Batuhan~Sariturk,~\IEEEmembership{Huawei Turkiye R\&D Center, }
      Rabia~Bayraktar,~\IEEEmembership{Huawei Turkiye R\&D Center, }\\
      Merve Elmas Erdem,~\IEEEmembership {Huawei Turkiye R\&D Center}
}

\maketitle

\begin{abstract}
With the rise of online education platforms, there is a growing abundance of educational content across various domain. It can be difficult to navigate the numerous available resources to find the most suitable training, especially in domains that include many interconnected areas, such as ICT. In this study, we propose a domain-specific chatbot application that requires limited resources, utilizing versions of the Phi language model to help learners with educational content. In the proposed method, Phi-2 and Phi-3 models were fine-tuned using QLoRA. The data required for fine-tuning was obtained from the Huawei Talent Platform, where courses are available at different levels of expertise in the field of computer science.
RAG system was used to support the model, which was fine-tuned by 500 Q\&A pairs. Additionally, a total of 420 Q\&A pairs of content were extracted from different formats such as JSON, PPT, and DOC to create a vector database to be used in the RAG system. By using the fine-tuned model and RAG approach together, chatbots with different competencies were obtained. The questions and answers asked to the generated chatbots were saved separately and evaluated using ROUGE, BERTScore, METEOR, and BLEU metrics. The precision value of the Phi-2 model supported by RAG was 0.84 and the F1 score was 0.82. 
In addition to a total of 13 different evaluation metrics in 4 different categories, the answers of each model were compared with the created content and the most appropriate method was selected for real-life applications. 
\end{abstract}


\IEEEpeerreviewmaketitle
\section{Introduction}

\IEEEPARstart{L}{arge} language models (LLMs) can easily solve many problems by leveraging their wide world knowledge during the pretraining phase. These multitask base models could easily be adapted to subtasks such as generation, translation, question-answering (Q\&A), and summarization. One of the most popular of these subtasks is Q\&A systems, namely chatbots. With the advent of GPT \cite{gpt2018}, we have integrated open-domain Q\&A systems into our daily lives. Over the years, several open-source LLMs have been released also, including T5 \cite{t5}, Llama \cite{touvron2023llama}, Phi \cite{li2023textbooksneediiphi15}, and Mistral \cite{jiang2023mistral}. There are different type of use cases for Q\&A systems. For example, search engines, intelligent customer services, and chatbots are emerging as open-domain systems recently \cite{chang2023survey}.

The improvement of LLMs provide opportunities to change the way perform many tasks. For example, ChatGPT became a very popular tool in daily life and is replacing search engines with additional features through the prompt tuning abilities. The adaptability of LLMs makes them suitable tools for improving efficiency of current applications. Chatbots are the best examples of this kind of usage as they can be adapted to many domains and could assist for an existing know-how more efficient. 

The reason behind the popularity of LLMs is their capacity of learned content and easy adaptation ability to other tasks because these models mostly have millions of parameters and this brings a huge complexity cost also and makes training and adaptation very hard. However, there are many cost effective methods that can be used for model alignment. For the encoder only models such as BERT, knowledge distillation methods could be used for training a smaller student model from a teacher model which mostly is a large foundation model. These small student models are usually used for inference. Quantization methods are also useful especially for faster inference and can be implemented after training by using methods such as Post-Training Quantization (PTQ) or before training by reducing unnecessary parameters using pruning methods. In recent years, the importance of small language models (SLMs) has increased with the development of models with fewer parameters but very effective in terms of accuracy, such as mistral and phi. These models are more accessible, cost-effective and shows comparable performances to LLMs  for various tasks. \par
In this study, a novel approach that combines the positive aspects of both  Retrieval Augmented Generation (RAG) and LLMs has been proposed to generate a Q\&A system. Performances of fine-tuned models, RAG systems that use base LLMs, and RAG systems that use fine-tuned LLMs were compared in a Q\&A manner. As LLMs, novel state-of-the-art Phi-2 and Phi-3 models have been selected to be used. These models attract attention with their remarkable success on benchmark datasets. Using Huawei Talent ICT Catalog documents and Q\&A pairs obtained from them, Phi-2 and Phi-3 Mini Instruct models were fine-tuned using Quantized Low-Rank Adaptation (QLoRA). In addition, RAG systems based on both these fine-tuned models and non-fine-tuned base models were generated. All these approaches were evaluated by calculating BLEU, ROUGE, METEOR, and BERTScore metrics, which are frequently used to evaluate Q\&A systems in the literature.
\section{Related Work}

\subsection{Question-Answering Systems} 

Developing Q\&A systems is challenging due to various complexities. Pretraining large parametric models have benefits for generalization, but it can also lead to non-factual responses and make it difficult for the model to efficiently retrieve necessary information. This situation, also known as hallucinating, is a common issue for open-domain Q\&A systems. Another challenge is ensuring that responses are ethical and preventing toxicity. This requires detailed evaluation and alignment of the models and most of the time requires human feedback. This involves detailed evaluation and alignment of the models, often requiring human feedback as well. Another problem may arise while adapting LLMs to close-domain use cases, in other words feeding new data to the model or adapting it to a specific domain that is not included in the training phase. In this case, models are more prone to hallucinations and it becomes more difficult to produce factual responses.

Our study is an example of this usage and differs from open-domain Q\&A systems because we expect our system to answer questions about specific know-how and generate factual results, but not to hallucinate. Close-domain Q\&A problems involve handling queries that require knowledge beyond what a language model is trained with \cite{clinical-qa,legal-qa}. Dealing with new data or retrieving information from a parametric model usually causes hallucinations \cite{lit6-survey}. However, RAG shows significant performances as a solution to this problem.

An example of using RAG for improving Q\&A systems is proposed in \cite{ft-rag}. The study advances the field by enabling the end-to-end training of the entire RAG architecture, which incorporates the Dense Passage Retrieval (DPR) \cite{dpr-lit2}. Utilizing pretrained BERT \cite{devlin2018bert} and Bidirectional and Auto-Regressive Transformers (BART) \cite{lewis2019bart} approaches, their methodology differs from prior implementations by training all components simultaneously. DPR, integral to the RAG, uses BERT models for encoding both questions and documents, while BART assists in the generation process. Updating the indexed knowledge base efficiently during training presents a challenge for this method. However it shows, enhancements in data encoding and re-indexing could improve the model's performance on Q\&A tasks.
 
Another specific teaching assistant use case proposed in \cite{ai-ta}, introduces an Artificial Intelligence (AI) teaching assistant architecture employing open-source LLMs, specifically LLaMA-2. The system integrates RAG, Supervised Fine-Tuning (SFT), and Direct Preference Optimization (DPO) to enhance the quality of answers on educational platforms. This implementation improved answer quality by 30\% and addressed scalability and data privacy in educational Q\&A systems.

\subsection{Fine-tuning, Large Language Models (LLMs)} 

As LLMs become popular, many applications have been developed where chatbots are used for different purposes. In addition to customized chatbots used in education platforms \cite{dhivvya2024buddybot}, models developed for learning and understanding a language that can be fine-tuned when necessary \cite{chen2024llama}, domain-specific models \cite{jeong2024fine}, and LLMs customized for emotion recognition \cite{peng2024customising} are examples of these.

As it is known, generalized language models trained with large datasets can be fine-tuned using task-specific data. General-purpose models, frequently used in the literature, can be customized and used for a specific task when necessary. The most commonly used method in the fine-tuning process of models is Low-Rank Adaptation (LoRA) \cite{lora}  which ensures memory efficiency by reducing trainable parameters. 

LoRA method was created to reduce model costs for many application areas where processing power and memory costs are critical. This structure aims to reduce the number of parameters and training time by using only a small number of parameters from each layer while freezing others. LoRA could be used for many domain-specific application such as cybersecurity awareness \cite{lora1}, task-agnostic, and task-specific \cite{lora2} models.

Due to their huge computational costs, optimizing LLMs is particularly important to ensure efficient performance. For example, a model with parameter size 7B needs at least 28GB of space and many models use float32 data type. Using different data types such as float16, float8 or integer can reduce computational complexity with a cost of lower accuracy. There are various quantization methods for reducing memory footprint of the models. For example, \cite{dettmers2024qlora} proposes an efficient usage of int8 conversion for quantization. Also there are libraries such as bitsandbytes \cite{bitsandbytes} specialized for this type of conversions that could be used with LLMs. These methods are used to prevent quantization loss to tensor multiplication and to find the optimum value. During the quantization process, inference time increases, but memory usage can be reduced by up to 71\% \cite{dettmers2024qlora, j2024finetuningllmenterprise}.

\subsection{Retrieval Augmented Generation (RAG)} 

Information retrieval in LLMs is challenging because learned information is stored implicitly in the model parameters for most of the pretrained LLMs \cite{devlin2018bert, liu2019roberta, t5}. This makes LLMs hard to adapt for generating factual responses in Q\&A systems. RAG approaches are used to overcome this problem and improve language models by easy alignment for various specific tasks.

The idea of using a non-parametric memory for retrieving external data for improving parametric language models is the starting point for developing RAG. Early studies generally use the retrieval model in the fine-tuning stage and train both retriever and generator models \cite{atlas}. Vector space representations also evolved in time in retrieval-augmented language models. Methods such as TF-IDF and BM25 are examples of strong sparse vector space representations. \cite{dpr-lit2} shows that DPR could outperform these traditional methods for Q\&A systems. It could also be used as a retriever component to provide latent documents that are conditioned on the input. DPR method is also used with small datasets in \cite{lewis2021rag}. 
\cite{lewis2021rag} uses BLEU and ROUGE metrics for the evaluation. Human evaluation is also very important for evaluating and preparing this kind of system for customer interaction and production. A custom human evaluation interface is also proposed in this study.

\cite{guu2020realm} is another study that proposes a retriever model evaluated for open Q\&A task. It is a latent knowledge retriever model that supports unsupervised training by explicitly exposing the role of world knowledge by asking the model to decide, unlike traditional language models. 
Retrieval-augmented architectures could also be used to improve model alignment with few-shot learning without increasing the parameter size. \cite{atlas} is an example of these kinds of studies which shows that jointly pretraining the retriever model and language model together improves the few-shot learning performance. RALM systems also could be used as in-context learning and could improve language model performances by keeping model parameters unchanged and without any further training \cite{ram2023incontextralm}.

\section{Materials and Method}
\subsection {Dataset}
\subsubsection{Q\&A pairs}

To fine-tune the models, course content that is available as open source on the Huawei Talent Platform ~\cite{huawei_ict_courses} was used. These courses are prepared at different levels and different domains such as AI, cloud, big data, or 5G.  

In this study, 500 Q\&A pairs were extracted from 14 courses for fine-tuning. Seven of these courses are general, and the remainder are professional courses. When generating question-answering pairs, we consider variations in sample sizes to ensure the model's generalization. Multiple Q\&A pairs, each covering different aspects of the same course, were added to the dataset, thus the LLM was expected to make more accurate inferences about the courses. We use GPT-4 \cite{achiam2023gpt} for generating question-answer pairs, and through human evaluation, we ensure the comprehensive inclusion of all specific course content details in addition to general knowledge across various domains and expertise categories. The generated Q\&A samples are shown in Table \ref{q&a}.

\renewcommand{\arraystretch}{1.5} 
\begin{table}[h!]
\centering
\caption{Example of Q\&A pairs}
\label{q&a}
\resizebox{\columnwidth}{!}{%
\begin{tabular}{p{5cm} p{7cm}}
\textbf{QUESTION} & \textbf{ANSWER}\\

\hline
What is the Huawei Talent System? &
  The Huawei Talent System refers to Huawei's comprehensive approach to nurturing, developing, and retaining talent within the company. \\
\hline
What are the key differences between HCIA, HCIP, and HCIE certifications in Huawei's certification system? &
  HCIA focuses on foundational knowledge, HCIP delves deeper into specific technologies, and HCIE demonstrates expert-level mastery in Huawei technologies. \\
\hline
What topics are covered in the general course on Introduction to 5G? &
  The general course on Introduction to 5G covers topics such as the evolution of mobile networks, key features of 5G technology, 5G architecture, deployment scenarios, and the impact of 5G on various industries. \\
\hline
Who is the target audience for the Search and Artificial Intelligence course? &
  The course is tailored for secondary specialized school students, junior college students, undergraduates, and general audiences with majors not related to information sciences or computers. It aims to make AI and search algorithms accessible to learners from diverse academic backgrounds. \\
\hline
How do students benefit from exploring commonly used algorithms in machine learning during the Foundations of AI module? &
  Exploring commonly used algorithms in machine learning during the Foundations of AI module allows students to gain insights into various machine learning techniques and their applications. By studying algorithms such as linear regression, decision trees, and support vector machines, students learn how to apply machine learning algorithms to solve real-world problems and make data-driven decisions.
\end{tabular}%
}
\end{table}

\subsubsection{PDF/PPT Information Extraction} 

The generated RAG system and the fine-tuned LLM aim to understand the courses on the Huawei Talent Platform, their contents, their relationships with each other. Accordingly, they will answer and guide the user's questions. To achieve this, Q\&A datasets have been generated using Huawei Course ICT Catalog PDF and PPT documents and GPT-4 model \cite{wolf-etal-2020-transformers}. Both files include information about the context of the courses, and additionally in the PDF file, there are tables that include information regarding the technical directions related to the courses, their names, versions, release dates, languages they are offered in and will be offered in the future.
Using GPT-4 to extract information from the documents, 20 Q\&A datasets were generated for each course, each containing ten Q\&A pairs, formatted in JSON. Furthermore, using the information from the “Corresponding Reference Table”, three additional datasets, each containing 50 pairs, have also been generated. A part of the relevant table can be seen in Table ~\ref{ict_academy}.

\begin{table}[h!]
\centering
\caption{Examples from the reference table of Huawei ICT Academy courses}
\label{ict_academy}
\renewcommand{\arraystretch}{1} 
\begin{tabular}{|>{\centering\arraybackslash}m{1.3cm}|>{\centering\arraybackslash}m{1.4cm}|>{\centering\arraybackslash}m{0.95cm}|>{\centering\arraybackslash}m{1.3cm}|>{\centering\arraybackslash}m{1.4cm}|}
\hline
\textbf{Technical Direction} & \textbf{Course Name} & \textbf{Version} & \textbf{Course Type} & \textbf{Available Languages} \\
\hline
Datacom & HCIA-Datacom & V1.0 & Certification Course & Chinese, English, French, Arabic, Portuguese, Spanish, German\\
\hline
AI & Deep Learning & V1.0 & Professional Course & Chinese \\
\hline
Cloud Computing & HCIA-Cloud Computing & V5.0 & Certification Course & Chinese, English, Spanish \\
\hline
5G & 5G Basics: What It's All About & V1.0 & General Course & English, Portuguese \\
\hline
\end{tabular}
\end{table}

\begin{figure}[h!]
  \begin{center}
  \includegraphics[width=1.0\linewidth]{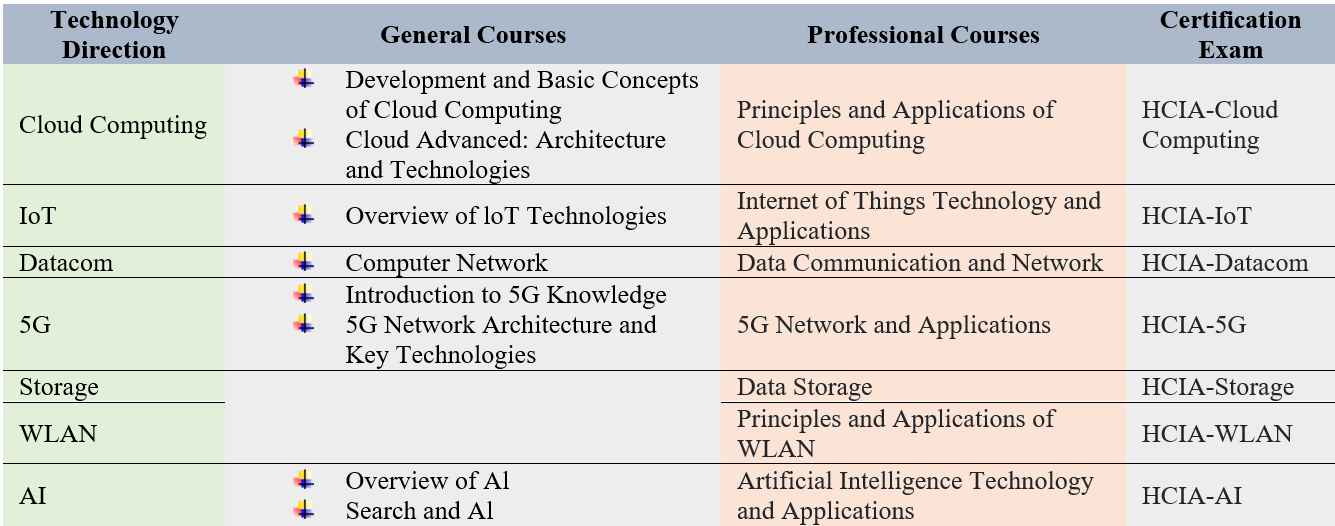}\\
  \caption{Path for study and exam}\label{study&exam}
  \end{center}
\end{figure}

Lastly, using the “Path for Study and Exam” table, which includes the technical directions covered by the courses, their relationships with each other and with related certification exams, another dataset containing 70 Q\&A pairs has also been generated. Related table is shown in Fig.~\ref{study&exam}.

\begin{figure}[h!]
    \centering
    \includegraphics[width=1.0\linewidth]{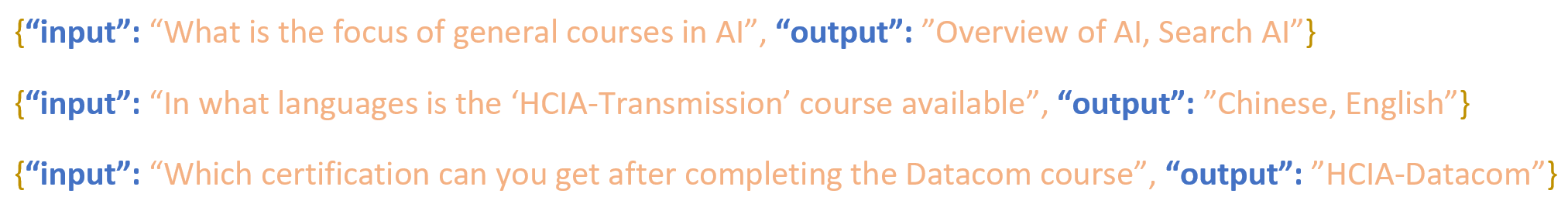}
    \caption{Q\&A pair examples from generated datasets}
    \label{pairs}
\end{figure}

Using detailed information about a total of 20 courses found in the Huawei Course ICT Catalog PDF and PPT documents, datasets containing a total of 420 Q\&A pairs have been generated to establish RAG systems. Example pairs from the generated datasets can be seen in Fig.~\ref{pairs}.

\subsection {Parameter Efficient Fine-Tuning}

Quantization is a method of expressing information with fewer bits. There are two basic stages during this conversion process. The first step is to find the normalization constant and scale the vector to the target range. The second step is the rounding process to the nearest degree to eliminate outlier values due to tensor multiplication. Elimination of outlier values is an important step to prevent quantization loss \cite{dettmers2024qlora}. If the outlier values are not used correctly in the tensor, the int8 conversion for a 32-bit Floating Point (FP32) tensor between [-127, 127] can be expressed as follows:

\small
\begin{equation}\label{eq1}
    X ^ {int8} = round \frac{127}{absmax({X^{FP32}})} =round ({c^{FP32}X^{FP32}})
\end{equation}
\normalsize where $c$ is the quantization constant or quantization scale. Dequantization is the inverse:

\begin{equation}\label{eq2}
    dequant(c^{FP32}, X^{int8}) =  \frac {X^{int8}}{c^{FP32}} = X^{FP32}
\end{equation} \par

LoRA is a fine-tune technique used to improve and expand the network layer of the model in which it is used, without having to change the model structure. The main purpose of LoRA is to add a layer of trainable weights without changing the original parameters of the network. To update LoRA's loss functions, Stochastic Gradient Descent (SGD) is adapted to the weights of the pretrained model. LoRA, which is generally injected into the linear layers of the model, is used between layers as follows \cite{dettmers2024qlora}:

\begin{equation}\label{eq3}
    XW=Y, 
    X\in \mathbb{R}^{b \times h}, 
    W\in \mathbb{R}^{h \times o}
\end{equation}

\begin{equation}\label{eq4}
    Y = XW + sXL_{1} L_{2}
\end{equation} Where $L_1 \in \mathbb{R}^{h \times r}$ and $L_2 \in \mathbb{R}^{r \times o}$ and $s$ is a scalar. \par

\begin{figure}[h!]
  \begin{center}
  \includegraphics[width=2.8in]{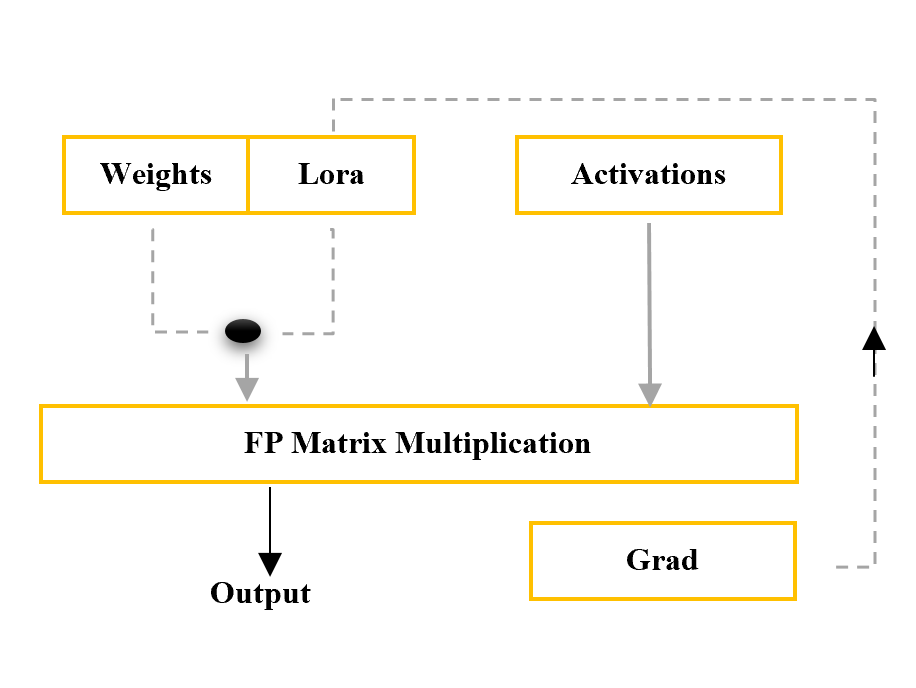}
  \caption{{A Simple Representation of QLoRA}}\label{qlora}
  \end{center}
\end{figure}

QLoRA (Quantized Low-Ranking Adaptation) ~\cite{dettmers2024qlora}, scales the weight values of the original network from float32, a high-resolution data type, to int4, a lower-resolution data type. This scaling process reduces memory needs and makes calculations faster. QLoRA is one of the most effective Parameter Efficient Fine Tuning (PEFT) techniques known. A representation of QLoRA is shown in Fig. ~\ref{qlora}.

QLoRA was developed to reduce memory usage; it can be used with 4-bit Normal Float, Double Quantization, and Paged optimizer techniques. 4-bit Normal Float can achieve 16-bit performance with 4-bit data with an equal number of elements assigned to the input tensor ~\cite{dettmers2024qlora, rangan2024fine}.

Phi-2 is a 2.7 million parameter SLM introduced by Microsoft. Phi-2 can understand text and code, produce customized creative content, and answer questions. Developed by Microsoft's NLP (Natural Language Processing) research team, Phi-2 was trained on a large text and code dataset and is reported to perform well on many tasks. The biggest advantage of Phi-2, whose potential applications are education, customer service, translation and creativity, is that it works more efficiently and faster than big data processing models with its fewer parameters \cite{egashira2024exploiting} \cite{abdin2024phi}.
Phi-3 is an SLM developed by Microsoft and offered in mini and medium versions. There are 4k and 128k instruction options for both mini and medium models. While the Phi-3-Mini model uses 3.8B parameters, the Phi-3-Medium model uses 14B parameters. A total of 3.3 trillion tokens obtained from synthetic data and data from public websites were used in training the Phi-3 model \cite{abdin2024phi}.

\subsection {Retrieval-Augmented Generation}
RAG is a tool that enhances the capabilities of LLMs. It does this by providing LLMs with access to external data sources \cite{yu2024evaluation}. This allows LLMs to generate responses that are enriched with this additional context. The additional context could be derived from a variety of sources, such as recent news articles or the audio transcripts from a lecture. Essentially, RAG can be thought of as an LLM that has been augmented with a vector search feature. Generally, a RAG system tackles tasks in a two-stage approach: First, it finds and retrieves relevant documents and constructs a task-specific prompt, then uses the generated prompt to create a response through a generator.\par
\begin{figure}[h!]
    \centering
    \includegraphics[width=0.6\linewidth]{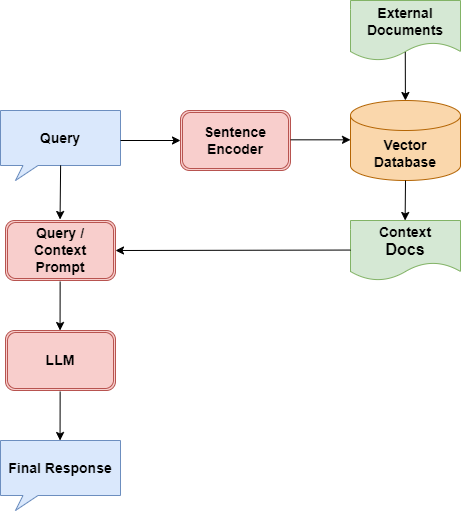}
    \caption{General structure of a RAG system}
    \label{rag}
\end{figure}

In an effort to address the challenges faced by LLMs, we introduce a chatbot that utilizes the RAG method. This approach allows the chatbot to search connected external sources containing information about Huawei ICT Academy courses and their relationships, and provide answers based on the retrieved guidelines. General structure and components of a RAG system are shown in Fig. \ref{rag}.

\subsection {RAG vs Fine-tuning}

\begin{table*}[htp]
\centering
\caption{RAG vs Fine-tune}
\small 
\begin{tabular}{|
p{0.1\linewidth}|
p{0.27\linewidth}| 
p{0.27\linewidth}| 
p{0.27\linewidth}|}
\hline
\textbf{Feature}&
\textbf{RAG}&
\textbf{Fine-tune}&
\textbf{RAG + Fine-tune}
\\
\hline
{\footnotesize \textbf{Timeliness}}&
{\footnotesize Can retrieve up-to-date information from external sources.}& {\footnotesize The model is dependent on the training data; therefore, it is as current as the data on which it is trained.} &
{\footnotesize Provides up-to-date and task-specific information with the ability to pull information from external sources and customized information.}\\
\hline
{\footnotesize \textbf{Efficiency}}&
{\footnotesize Instead of keeping large data sets in the model's memory, it pulls data from external sources when necessary.}& 
{\footnotesize A model optimized for a particular task can perform more efficiently.} &
{\footnotesize While it is optimized for specific tasks, it can use external sources of information when necessary.}\\
\hline
{\footnotesize \textbf{Flexibility}}&
{\footnotesize Can access different and diverse sources of information.}& 
{\footnotesize It is necessary to master the training process and training data.} &
{\footnotesize It can be integrated into different information sources and retrained for domain-specific studies.}\\
\hline
{\footnotesize \textbf{Variety}}&
{\footnotesize It can access information that is not in the model's training data.}& 
{\footnotesize It uses data customized for a specific task.} &
{\footnotesize When using the information in the customized data set, it can draw additional information from external sources.}\\
\hline
{\footnotesize \textbf{Dependency}}&
{\footnotesize It is directly dependent on external sources to retrieve information.}& 
{\footnotesize Retraining is required for every new information, regardless of external sources.} &
{\footnotesize It is dependent on both external resources and the fine-tuning process. In this way, it is aimed to generate more reliable responses.}\\
\hline
\end{tabular}
\label{rag_vs_finetune}
\end{table*}

RAGs enable a trained model to generate a response by retrieving data from different contents to establish its relationship with the model-content. After training a model, including new data requires time and human resources. Using RAG, new data can be pulled from a vector database without retraining the model.
Also, trying to keep a large amount of data in the model's memory while the model is being trained will increase both time and resource usage. RAG reduces cost while increasing model efficiency by using it when necessary. Besides these, there are some disadvantages such as dependence on external sources in the process of retrieving information from a database, delay in information retrieval and increased complexity. To overcome these disadvantages, a new approach is needed to reduce the response time and complexity of the model. \par
The model used in RAGs can be fine-tuned to a domain-specific area in a short time and the model that will emerge as a result of fine-tuning requires as few resources as possible. As it is known, the fine-tuning process allows the model to perform better on a domain-specific task and does not depend on external information sources when generating a response. However, it has disadvantages such as timeliness and time efficiency. 
So, a new method can be proposed by combining the separate advantages of RAG and fine-tuning. This method can be a RAG system in which SLMs can be fine-tuned in a short time by using QLora and vector databases can ensure that the model is constantly updated are actively used. The advantages we foresee when using the fine-tuned model with RAG are detailed in Table \ref{rag_vs_finetune}.\par

\subsection {Proposed Method}

In this study, we aim to create a customized chatbot to help learners select a learning path to specialize in a domain. In our case, the domain is one of the ICT-related fields, such as cloud, AI, etc. The performance of different alignment methods, given limited data and computational resources, was analyzed using Phi-2, Phi-3, and fine-tuned versions of these models.

Our pipeline includes three main steps: generating a Q\&A pair dataset for fine-tuning, parameter efficient fine-tuning of models and using RAG method to improve efficiency using course catalogs as external sources for generating vector databases.

\begin{figure}[h!]
    \centering
    \includegraphics[width=1.0\linewidth]{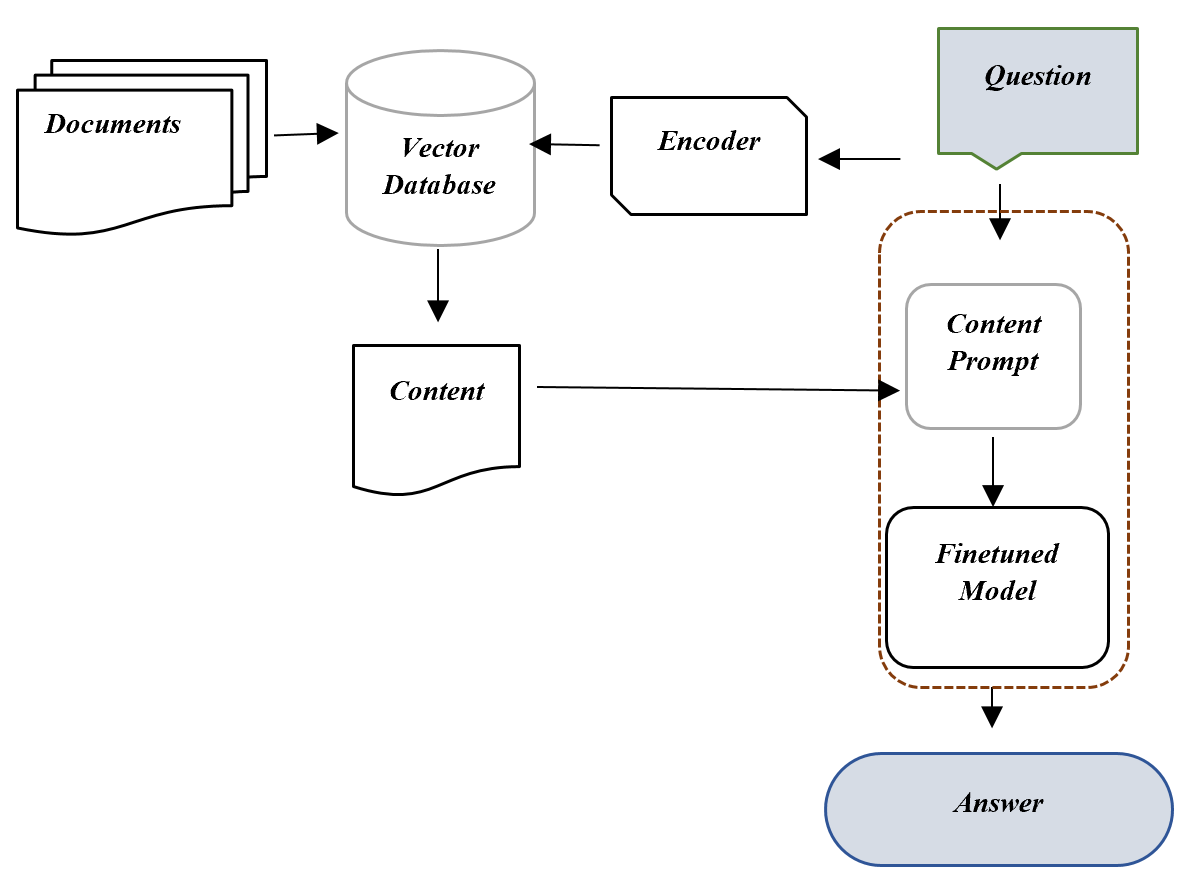}
    \caption{Proposed Method}
    \label{meraba}
\end{figure}

Fine-tuning phase was carried out with parameter efficient methods to reduce cost complexity. QLoRA method utilized here for both models. 500 Q\&A generated pairs are used for model fine-tuning. RAG method implemented using external sources in different forms such as PPT and DOC to create vector databases. We also utilized pretrained retrieval model for this part. General structure of our proposed method is shown in Fig. \ref{meraba}



\subsection{Evaluation Metrics}

To evaluate the performance of fine-tuned, RAG, and RAG+fine-tuned approaches, several commonly used methods from the literature were selected. The chosen metrics were; BLEU ~\cite{papineni-etal-2002-bleu}, ROUGE ~\cite{lin-2004-rouge}, METEOR ~\cite{banerjee-lavie-2005-meteor}, and BERT ~\cite{devlin2018bert} score. \par
Each of these metrics evaluates different aspects of the systems. For example; BLEU is particularly valuable for assessing the context relevance, faithfulness, and answer relevance of the generated text. ROUGE, on the other hand, is crucial for evaluating how comprehensively the generated text captures the essential information contained in the reference text, making it essential for completeness assessment. METEOR extends this by evaluating not only the accuracy and completeness, but also the fluency and paraphrasing quality of the generated text, considering synonyms and grammatical variations. Lastly, BERTScore provides a deep semantic evaluation by analyzing the contextual alignment of the generated text with the reference, ensuring that the responses are not only correct in detail but also right in essence and context. Together, these metrics provide a holistic view of system performance across various dimensions critical to the success of used systems in generating relevant and accurate answers. \par
BLEU is a popular metric used to evaluate the quality of text generated by machine translation systems against a set of high-quality reference translations. It was developed by IBM researchers in 2002 \cite{papineni-etal-2002-bleu} and has been extensively applied not only in machine translation but also in other NLP tasks, such as automated Q\&A systems. BLEU assesses the quality of the generated text by measuring the overlap of n-grams, which are contiguous sequences of 'n' words, between the machine-generated output and the reference text. Then, scores are combined using geometric mean and multiplied by an exponential Brevity Penalty (BP) factor to account for overly short translations. BLEU score ranges from 0 to 1, where 1 denotes a perfect match with the reference text. Despite its widespread use due to its strong correlation with human judgments, it is important to note that BLEU primarily measures lexical similarity and may not fully capture semantic accuracy or fluency in the generated text. In BP, c is the length of the generated output and r is closest/best match reference length. 

\begin{equation}\label{eq5}
    BP = 
    \begin{cases} 
    1 & \text{if } c > r \\
    e^{(1-r/c)} & \text{if } c < r 
    \end{cases}
\end{equation}

\begin{equation}\label{eq6}
    BLEU=BP\cdot e^ {(\sum _ {n=1}^ {4} \frac {1}{4}\log {P}_n)}
\end{equation}

ROUGE, developed by Chin-Yew Lin \cite{lin-2004-rouge}, is an evaluation metric extensively used to assess the quality of summaries produced by automatic text summarization systems and is highly effective for evaluating text generation tasks like Q\&A assistants. This metric emphasizes recall, focusing on the extent to which the generated text captures the content found in the reference text. ROUGE measures the quality of text by comparing it to a set of reference summaries, assessing n-gram overlaps with variants such as ROUGE-1 and ROUGE-2 for 1-gram and 2-gram overlaps, respectively, to measure the content accuracy. Additionally, ROUGE-L analyzes the longest common sub-sequence to evaluate the fluency and order of content, while ROUGE-Lsum extends this analysis to the entire text, offering insights into structural and sequential similarity. Although ROUGE, similar to BLEU, effectively measures n-gram overlap and doesn't account for semantic similarity or answer correctness, it provides a nuanced evaluation framework that enables a detailed assessment of how well a system reproduces relevant answers and maintains logical coherence in its responses.

\small
\begin{equation}\label{eq7}
\text{ROUGE-N} = \frac{\sum_{S \in \text{Reference Sum.}} \sum_{\text{gram}_n \in S} \text{Count}_{\text{match}}(\text{gram}_n)}{\sum_{S \in \text{Reference Sum.}} \sum_{\text{gram}_n \in S} \text{Count}(\text{gram}_n)}  
\end{equation}
\normalsize

\begin{equation}\label{eq8}
\text{ROUGE-L} = \frac{\sum_{S \in \text{Reference Sum.}} \text{LCS}(\text{Candidate}, S)}{\sum_{S \in \text{Reference Sum.}} \text{Length}(S)}  
\end{equation}

METEOR is a metric designed to overcome some of the limitations of earlier metrics, by incorporating a more sophisticated analysis. Developed by researchers at Carnegie Mellon University \cite{banerjee-lavie-2005-meteor}, METEOR provides a nuanced analysis of the generated output by comparing it not only on exact word correspondences but also on synonyms, morphological variations, and paraphrases. This allows it to assess semantic and morphological similarities between the generated and reference outputs more comprehensively.\par
METEOR operates by aligning words between the generated and reference texts through a variety of match types, including exact, synonym, and paraphrase matches. It calculates scores based on these alignments, using a harmonic mean of precision and recall, with a stronger emphasis on recall. This balance can be adjusted to favor different aspects of the answers. Additionally, the metric incorporates a penalty for too many short matches, encouraging answers that are not only accurate but also fluently structured. This enhanced capability to recognize semantic equivalents makes METEOR particularly valuable for evaluating Q\&A systems, where the quality of an answer is not solely dependent on textual similarity but also on the preservation of meaning and context from the question.

\begin{equation}\label{eq9}
P = \frac{\text{number of matched unigrams in candidate}}{\text{total number of unigrams in candidate}}
\end{equation}

\begin{equation}\label{eq10}
R = \frac{\text{number of matched unigrams in candidate}}{\text{total number of unigrams in reference}}
\end{equation}

\begin{equation}\label{eq11}
\text{Penalty} = 0.5 \left( \frac{\text{number of chunks}}{\text{number of unigram matches}} \right)^3
\end{equation}

\begin{equation}\label{eq12}
F_\text{mean} = \frac{10 \times P \times R}{R + 9 \times P}
\end{equation}

BERTScore \cite{zhang2019bertscore} is an advanced evaluation metric that utilizes the deep contextual embeddings from the BERT model to assess the quality of generated text. Unlike traditional metrics that rely on surface-level word matches, BERTScore computes the semantic similarity between the generated and reference texts by measuring the cosine similarity of their BERT-generated embeddings. This method allows BERTScore to capture the contextual nuances more effectively, by going beyond traditional n-gram matches to capture deeper linguistic and semantic nuances. The metric evaluates the performance based on precision, recall, and the F1 score, which are derived from the highest cosine similarities across token embeddings in both texts. Precision measures the relevance of each token in the generated text, recall assesses how well the reference text is captured, and the F1 score provides a balanced measure of both, making BERTScore particularly suitable for tasks where deep semantic understanding is crucial.

\section{Experiment and Results}
\subsection{Retrieval-Augmented Generation with Phi}

All experiments regarding the RAG system were carried out using Nvidia T4 GPU. We utilized the Transformers library from Hugging Face \cite{wolf-etal-2020-transformers} and used the state-of-the-art, open source LLMs, Phi-2 and Phi-3-Mini, to generate the RAG system. Additionally, LangChain library \cite{Chase_LangChain_2022} is used to create all the necessary components.\par
To optimize the models, we employed a technique known as model quantization. This process involves reducing the number of trainable parameters of a model by representing the values using lower precision data types, such as 8-bit integers, instead of high precision numbers, like 32-bit floats. This significantly reduces memory usage, speeds up model execution, and maintains accuracy at an acceptable level. We used the bitsandbytes library \cite{Tim_Bitsandbytes_2021} for this purpose, which works seamlessly with the Transformers library, simplifying the quantization process. Bitsandbytes is library particularly known for its 8-bit optimizers, quantization functions, and matrix multiplication capabilities for both 4-bit and 8-bit operations. By implementing model quantization, we significantly reduce the computational load, allowing for more efficient handling and retrieval of vector data. This seamless interaction between reduced model size and enhanced retrieval processes is essential for the effective functioning of the vector database within our RAG system.\par
The mentioned vector database, is one of the key components of a RAG system. This is a storage system that holds vector representations, or "embeddings", of the documents in the provided dataset. These embeddings are generated by transforming the text data into a numerical format that a machine learning model can understand. When a new user query comes in, the system transforms this query into its vector representation and then searches the vector database for the most similar vectors, a process known as "efficient retrieval". This allows for efficient and scalable retrieval of relevant documents, which is crucial for large-scale information retrieval tasks. We used Facebook AI Similarity Search (FAISS) to create the vector database \cite{douze2024faiss}. 

FAISS provides a collection of algorithms to quickly search for similar vectors in large databases, and renowned for its efficient and scalable similarity search capabilities. \par
To create this vector database, we converted the JSON files and the PDF file into raw text. We then chunked this text into manageable pieces and loaded these chunks into the FAISS index. This process allows us to search and find which parts of the documents are relevant, especially when dealing with large amounts of data.\par

\begin{figure}[h!]
    \centering
    \includegraphics[width=1\linewidth]{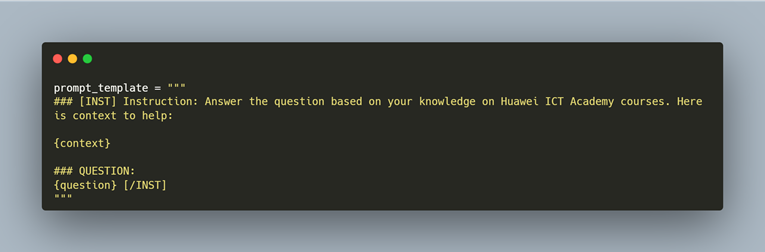}
    \caption{Used prompt template}
    \label{prompt_temp}
\end{figure}

After performing these steps, we need a “retriever”. The retriever acts as an intermediary that facilitates communication between the generated vector database and the LLM. As mentioned earlier, we used LangChain library to create the retriever and other necessary components. The "all-mpnet-base-v2" sentence transformer model is used to create vector embeddings and, as retriever settings, the “similarity” search type is employed with the top $k=5$ pairs with the highest similarity score.\par
Finally, we created chains for the Phi models, and updated the prompt template to allow contexts to be passed in Fig. \ref{prompt_temp}. These contexts are the ones that the FAISS index gathered. After integrating the LLM Chain with the FAISS retriever, we put them together to complete our RAG system.

\begin{table}[h!]
\centering
\caption{Comparison of Metrics for RAG Systems}
\label{metrics_comparison_2}
\renewcommand{\arraystretch}{1} 
\begin{tabular}{|>{\centering\arraybackslash}m{2.5cm}|>{\centering\arraybackslash}m{1.5cm}|>{\centering\arraybackslash}m{1.5cm}|>{\centering\arraybackslash}m{1.5cm}|}
\hline
\textbf{Metrics} & \textbf{Phi-2 (JSON)} & \textbf{Phi-3-Mini-4k Instruct (JSON)} & \textbf{Phi-3-Mini-4k Instruct (PDF)} \\
\hline
BLEU & 0.003 & 0.043 & \underline{0.051} \\
Unigram Precision & \underline{0.394} & 0.291 & 0.284 \\
Bigram Precision & \underline{0.084} & 0.065 & 0.070 \\
Trigram Precision & 0.025 & 0.021 & \underline{0.028} \\
4-gram Precision & 0.010 & 0.008 & \underline{0.012} \\
ROUGE-1 & \underline{0.324} & 0.315 & 0.297 \\
ROUGE-2 & \underline{0.107} & 0.088 & 0.073 \\
ROUGE-L & \underline{0.209} & 0.179 & 0.159 \\
ROUGE-LSum & \underline{0.265} & 0.245 & 0.221 \\
METEOR & 0.219 & 0.272 & \underline{0.274} \\
BERTScore-F1 & \underline{0.879} & 0.852 & 0.846 \\
BERTScore-Precision & 0.838 & \underline{0.852} & 0.843 \\
BERTScore-Recall & \underline{0.858} & 0.852 & 0.844 \\
\hline
\end{tabular}
\end{table}

After the RAG system was set up and fed with the generated datasets, questions that were formulated to test the models were used to obtain answers both from the pretrained models and from the RAG system. Answers to some selected example questions can be seen in Table \ref{rag_test}.

\begin{table*}[h!]
\centering
\caption{Example test question and answers}
\label{rag_test}
\small
\begin{tabular}{|>{\centering\arraybackslash}m{0.15\linewidth}|>{\raggedright\arraybackslash}m{0.8\linewidth}|}
\hline
\textbf{Question} & How do students gain hands-on experience in the 'Principles and Applications of WLAN' course? \\ \hline
\textbf{Ground Truth} & In the "Principles and Applications of WLAN" course, students gain hands-on experience through several practical approaches designed to enhance their understanding and skills in wireless local area networking. Here’s how this is typically achieved: Lab Sessions, Simulation Tools, Practical Projects, Case Studies, Interactive Demonstrations. \\ \hline
\textbf{Phi-2 JSON} & Students can gain hands-on experience in the 'Principles and Applications of WLAN' course through activities such as WLAN design projects, simulation exercises, and practical demonstrations where they get to configure and troubleshoot WLANs. \\ \hline
\textbf{\begin{tabular}[c]{@{}c@{}}Phi-3-Mini-4k \\ Instruct (JSON)\end{tabular}} & Students gain hands-on experience in the 'Principles and Applications of WLAN' course through several interactive activities such as WLAN design projects, simulation exercises, and practical demonstrations. These components allow learners to actively engage with the material by applying theoretical knowledge to real-world scenarios, thereby enhancing their understanding and proficiency in setting up, configuring, and troubleshooting Wireless Local Area Networks (WLANs) across different environments like homes, offices, and public spaces. In the 'Principles and Applications of WLAN' course, students can expect to participate in hands-on experiences that solidify their learning and provide them with practical skills. \\ \hline
\textbf{\begin{tabular}[c]{@{}c@{}}Phi-3-Mini-4k\\ Instruct (PDF)\end{tabular}} & Students can gain hands-on experience in the "Principles and Applications of Wireless Local Area Networks (WLAN)" course through various practical activities, labs, and projects that are typically integrated into the curriculum. These may include:  1. Lab sessions: Students often participate in lab exercises where they set up and configure wireless networks using different hardware devices such as routers, access points, and network analyzers. They learn how to measure signal strength, analyze data traffic, and troubleshoot common issues related to WLAN performance.  2. Project assignments: In some courses, students work on group or individual projects that require them to design, implement, and evaluate a real-world WLAN solution. This could involve setting up a small office network, creating an ad hoc network for testing purposes, or developing a mobile device connectivity system.  3. Case studies: The course might also incorporate case studies from industry professionals, which provide insights into real-life challenges faced by organizations when deploying and managing WLAN systems. Analyzing these cases helps students understand the complexities involved in WLAN deployment and management.  4. Simulation tools: Some courses use simulation software like NS2 or OPNET to model and simulate WLAN environments. \\ \hline
\end{tabular}
\end{table*}

To evaluate the performance of the RAG systems, the same evaluation metrics that used for fine-tuned models were also used. When the results obtained using evaluation metrics are examined, the RAG system uses Phi-3-Mini-4k base model as the LLM and the PDF file includes information about courses as the external data source has the highest BLEU and METEOR metric results among all three RAG systems, with 0.051 and 0.274, respectively. On the other hand, for ROUGE metric score and BERTScore, the system uses Phi-2 and generated JSON files achieved the highest results with 0.324 ROUGE-1, 0.107 ROUGE-2, 0.209 ROUGE-L, 0.265 ROUGE-Lsum scores, and 0.858 F1 score calculated using BERTScore. All metrics results for RAG systems are shown in Table ~\ref{metrics_comparison_2}. 

\subsection{Model Fine-Tuning}

To fine-tune the Phi-2 model; 2 hours of training was conducted using A100 GPU, 2 batch-size, 22 epoch, 5k steps, with 500 Q\&A pairs obtained from the Huawei talent platform. The questions asked to the fine-tuned model and their answers are shown in Table ~\ref{model_answers}.
In the table given, it can be seen that the answers provided by the chatbot are largely related to the Huawei talent platform. It was observed that semantic integrity could be achieved in the answers and a connection could be established between the question and the answer. As it is known, semantic integrity alone may not be sufficient to evaluate the chatbot's answers correctly.
For this reason, different evaluation metrics were used to evaluate the fine-tuned models. One of these metrics; BERTScore is a metric that calculates the similarity between the answer given by the chatbot to a question and the Ground Truth (GT) text. In the similarity calculation, it evaluates not only the number of matched words but also the semantic context. While the precision and recall values of the fine-tuned Phi-2 model are more than 0.80, the F1 score is 0.76. In short-term trained LLMs, where the scope is wide but the training data is low, keeping the BERTScore above 0.70 is important for model performance. \par

\begin{table}[b!]
\centering
\caption{Comparison of Metrics for Phi-2 and Phi-3}
\label{metrics_comparison_1}
\renewcommand{\arraystretch}{1} 
\begin{tabular}{|>{\centering\arraybackslash}m{3cm}|>{\centering\arraybackslash}m{1.25cm}|>
{\centering\arraybackslash}m{1.25cm}|>
{\centering\arraybackslash}m{1.5cm}|}
\hline
\textbf{Metrics} & \textbf{Phi-2} & \textbf{Phi-3} &\textbf{Phi-2+RAG}  \\ 
\hline
\text{BLEU} & 0.0187 & 0.0230 & \underline{0.0454}\\
\text{Unigram Precision} & \underline{0.4265} & 0.3728 & 0.3780\\
\text{Bigram Precision} & \underline{0.1038} & 0.0839 & 0.0959\\
\text{Trigram Precision} & 0.0341 & 0.0306 & \underline{0.0371}\\
\text{4-gram Precision} & 0.0161 & 0.0135 & \underline{0.0204}\\
\text{ROUGE-1} & 0.3013 & 0.2692 & \underline{0.3315}\\
\text{ROUGE-2} & 0.0917 & 0.0856 & \underline{0.1012}\\
\text{ROUGE-L} & 0.2006 & 0.1801 & \underline{0.2027} \\
\text{ROUGE-LSum} & 0.2409 & 0.2264 & \underline{0.2490} \\
\text{METEOR} & 0.2029 & 0.1693 & \underline{0.2349} \\
\text{BERTScore-F1} & 0.7628 & 0.7456 & \underline{0.8270} \\
\text{BERTScore-Precision} & 0.8168 & \underline{0.8711} & 0.8454\\
\text{BERTScore-Recall} & 0.8083 & 0.8281 & \underline{0.8434} \\
\hline
\end{tabular}
\end{table}

\begin{table*}[h!]
\centering
\caption{Example of Model Answers}
\small 
\begin{tabular}{|>{\raggedright\arraybackslash}p{0.2\linewidth}|>{\raggedright\arraybackslash}p{0.37\linewidth}|>{\raggedright\arraybackslash}p{0.35\linewidth}|} 
\hline
\textbf{Question}&\textbf{Phi-2}&\textbf{Phi-3}\\
\hline
{\footnotesize What are the distinguishing characteristics of HCIP, HCIA, and HCIE certifications within Huawei's certification system, and how do they cater to different levels of expertise in Huawei technologies?}
 & {\footnotesize HCIP offers advanced specialization, HCIA focuses on foundational knowledge, and HCIE demonstrates expert-level mastery in Huawei technologies. These certifications cater to diverse skill levels, ensuring accessibility for beginners while providing rigorous standards for experienced professionals.} &
 {\footnotesize HCIP, HCIA, and HCIE certifications are part of Huawei's certification system designed to validate the skills and knowledge of professionals working with Huawei technologies. They cater to different levels of expertise, with HCIP focusing on Huawei Certified Network Products, HCIA targeting a broader spectrum of Huawei technologies including software and solutions, and HCIE representing the highest level of expertise.}\\
\hline
{\footnotesize Describe the primary focus of the '5G Network and Applications' course.}&
{\footnotesize The primary focus of the '5G Network and Applications' course is to provide an in-depth understanding of the technical aspects, standards, and applications related to 5G networks. It covers topics such as network architecture, deployment models,core network technologies, and use cases for 5G technology.}&
{\footnotesize The '5G Network and Applications' course focuses on providing an understanding of 5G networks and applications, including exploration of key features and benefits of 5G technology.}\\
\hline
{\footnotesize Can you provide an overview of the role of the course satisfaction survey in the '5G Basics: What it's all about', '5G Network and Applications', and 'Introduction to Computer Networks and the Internet' courses?}&
{\footnotesize In the 'Introduction to 5G Knowledge' course, the satisfaction survey helps instructors understand students' understanding of key concepts, their level of engagement with the material, and suggestions for improving the course content. In the '5G Network and Applications' course, the survey assists instructors in evaluating the effectiveness of the curriculum and identifying areas for improvement. In the 'Introduction to Computer Networks and the Internet' course, the survey provides insights into students' comprehension of network principles and practical skills acquired throughout the course.}&
{\footnotesize In the '5G Basics: What it's all about' course, the course satisfaction survey serves as a feedback mechanism to evaluate students' understanding of the fundamental concepts of 5G technology. In the '5G Network and Applications' course, the course satisfaction survey evaluates students' satisfaction with the exploration of 5G network architectures, protocols, and applications. In the 'Introduction to Computer Networks and the Internet' course, the course satisfaction survey evaluates students' satisfaction with the integration of 5G technology into computer networks.}\\
\hline
{\footnotesize What types of skills will students gain upon completion of the 'Data Communication and Network' course?}&
{\footnotesize Upon completion, students will gain practical skills in configuring, networking, and troubleshooting enterprise networks using routers, switches, WLAN, and security technologies. They will also be proficient in performing routine network operations and maintenance tasks, as well as designing and implementing scalable and secure network solutions.}&
{\footnotesize Upon completion of the 'Data Communication and Network' course, students will gain skills in configuring and troubleshooting LAN and WAN networks, performing routine network maintenance, implementing security measures, and analyzing network performance.} \\
\hline
\end{tabular}
\label{model_answers}
\end{table*}

The number of words used in the questions and answers used for training in fine-tuned models varies. For the model to be generalize, it must be able to create meaningful sentences with different numbers of words. It has been observed that the fine-tuned Phi-2 model can give answers of different lengths to the questions asked. However, it would not be correct to evaluate only the length or shortness of the answers. Instead, the correspondence rate between the question and the answers can be examined.
ROUGE calculates the bi-gram and n-gram correspondence between ground truth (GT) and answers. This calculation evaluates how similar the answer is to the ground truth text at the word level. While making this calculation, normalization is applied to the texts to avoid high score differences that could arise between long and short texts.\par

In Table ~\ref{metrics_comparison_1}, the ROUGE value of the Phi-2 model is calculated as 0.30 for single-word groups, 0.09 for double-word groups and 0.20 for the longest common word group. We can decide which of these ratios we should consider according to our expectations from the model we train. If there is a limited resource, training time is short, and a study is being conducted with limited data, it would be more accurate to make an evaluation based on single-word groups. The ROUGE value obtained as 0.30 can be improved by adding to the dataset and extending the training time. \par

To fine-tune the Phi3-Mini-4k model with QLoRA, the same parameters as Phi-2 were used. Lora rank $32$, alpha $64$, dropout $0.05$ was used and Lora was applied to all linear layers. $2.5 . e^{-5}$ learning rate and 8-bit Adam optimizer were used in the $5k$ step trained model. \par

\begin{figure}[h!]
    \centering
    \includegraphics[width=1.0\linewidth]{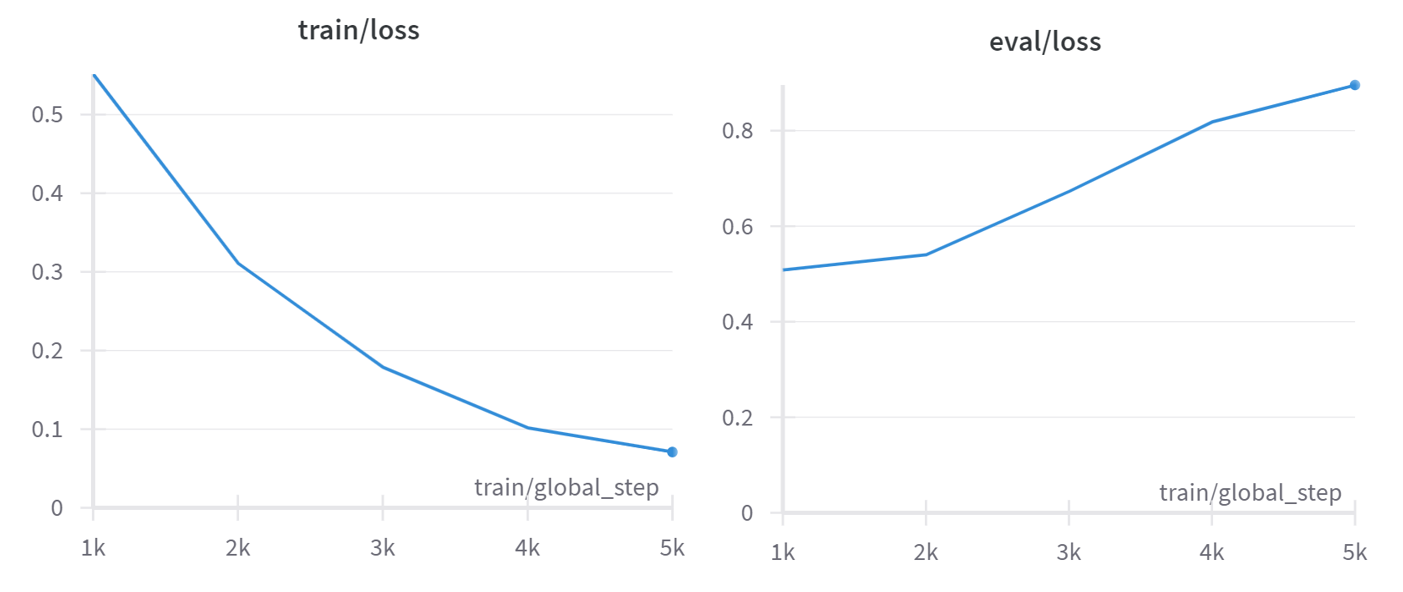}
    \caption{Phi-3 Train and Validation Loss Graphs}
    \label{phi3_loss}
\end{figure}

The Loss graphs given in Fig.~\ref{phi3_loss} were examined during the training and a gradual decrease in training loss was observed. In the trained model, the training loss decreased to 0.07. When working with limited data, training can be cut into $3k-4k$ steps to prevent the model from memorizing, and the number of steps can be completely rearranged according to the loss we expect from the model.

The fine-tuned Phi-3 model was asked the same questions as the Phi-2 model, and the answers are shown in Table ~\ref{model_answers}. As is known, Ground Truth (GT) answers to each question must be prepared to calculate the metrics shown in Table ~\ref{metrics_comparison_1}. Evaluation metrics can only be calculated correctly in this way based on a standard. In this study, GT answers prepared by adhering to the Huawei talent platform have also been added to Table I of Supplementary Material.\par
When the answers of the Ground Truth and Phi models are compared in terms of content, it is seen that the Phi-2 model can establish the content relationship better than Phi-3. \par
As can be seen from Table ~\ref{metrics_comparison_1}; in the Phi-2 model BLEU, ROUGE and METEOR metrics are slightly ahead. However, the Phi-3 model surpassed the Phi-2 model in terms of BERTScore-Precison and BERTScore-Recall. \par
To make a correct evaluation here, it is necessary to correctly define the means of the BERT score for this study. As it is known when calculating the BERT score, the number of matched words is as important as semantic relatedness. In Table I of Supplementary Material, it is seen that the answers of the Phi-3 model are longer than the Phi-2 model. A longer number of words used in the answers will positively increase the number of words matched, and this increase will also have a positive effect on the average scores.

\subsection{Proposed Method}

In the proposed method, RAG-supported chatbot experiments were carried out using the Phi-2 model, which gave the answers closest to what we expected from the fine-tuned models. By the methodology shown in Fig. ~\ref{meraba}, a vector database was first created with the documents obtained from the open source Huawei Talent Platform. The content extracted from the database and the content-prompt extracted from the question were combined and given as input to the fine-tuned model, and the model outputs were examined. An example of the content prompt used in the created pipeline is shown in Fig. ~\ref{prompt_temp}.

While creating the LLM chain, the tokenizer was created using the Phi-2 base model and the transformers text generation pipeline from the Python libraries was used. To create the text generation pipeline, the temperature was set as $0.2$, repetition penality as $1.1$ and max new tokens as $300$. These values may vary depending on our expectations from the created chatbot.
After the text generation pipeline was created, the Phi-2 model that we fine-tuned was used as a pretrained weight using the “PeftModel” library defined in Python. The questions given in Table I of Supplementary Material were asked to the fine-tuned model used with the vector database and the answers were recorded.\par

When the answers given in Table ~\ref {model_answers} and Table I of Suplementary Material are examined in terms of context; it is largely similar to the data obtained from the course catalogs. In addition, it was observed that the fine-tuned Phi-2 and Phi-3 models were able to establish better contextual relationships than the answers given by them. The evaluation metrics given in Table ~\ref{metrics_comparison_1} can be used to evaluate the number of matched words and semantic integrity.
Based on the evaluation metrics, the Phi-2 model supported by the vector database has surpassed other models in almost all metrics. It is seen that it has widened the gap with its competitors, especially in ROUGE-1 and METEOR metrics.
As it is known, ROUGE can be used to evaluate text summarization models, and models that extract content from the vector database created with RAG are examples of these. Therefore, leaving its competitors behind in this metric shows that it can successfully extract content from the vector database. The ROUGE-L score calculated as 0.20 shows that there is an increase in the number of matched words even in long texts.
In applications where METEOR translation is performed, it returns a value between 0 and 1, depending on the order and frequency of words. As a result of supporting the model we trained with a vector database, the METEOR value was increased from 0.20 to 0.23. This value can be further increased by expanding the dataset, adding new documents that can create content for the RAG pipeline, and optimizing the training parameters.

\subsection{Strength \& Weakness}

It is known that many LLMs mentioned in the literature have been trained for different purposes. More than one version of the same model may be offered. These versions can be used for their intended purpose in dialogue, summarization and translation-based applications. Among the known models; models such as Mistral \cite{jiang2023mistral}, Llama \cite{chen2024llama}, and Phi \cite{abdin2024phi} are also open-source accessible and can be used without token limits. The common feature of these models is that they are small, quantizable models trained on large datasets. In this study, unlike the studies \cite{Bayraktar2024} that discussed all these models with their negative and positive aspects, we evaluated the Phi models offered in different versions by the same developer.


\begin{figure*}[t!]
  \centering
  \includegraphics[width=0.80\textwidth]{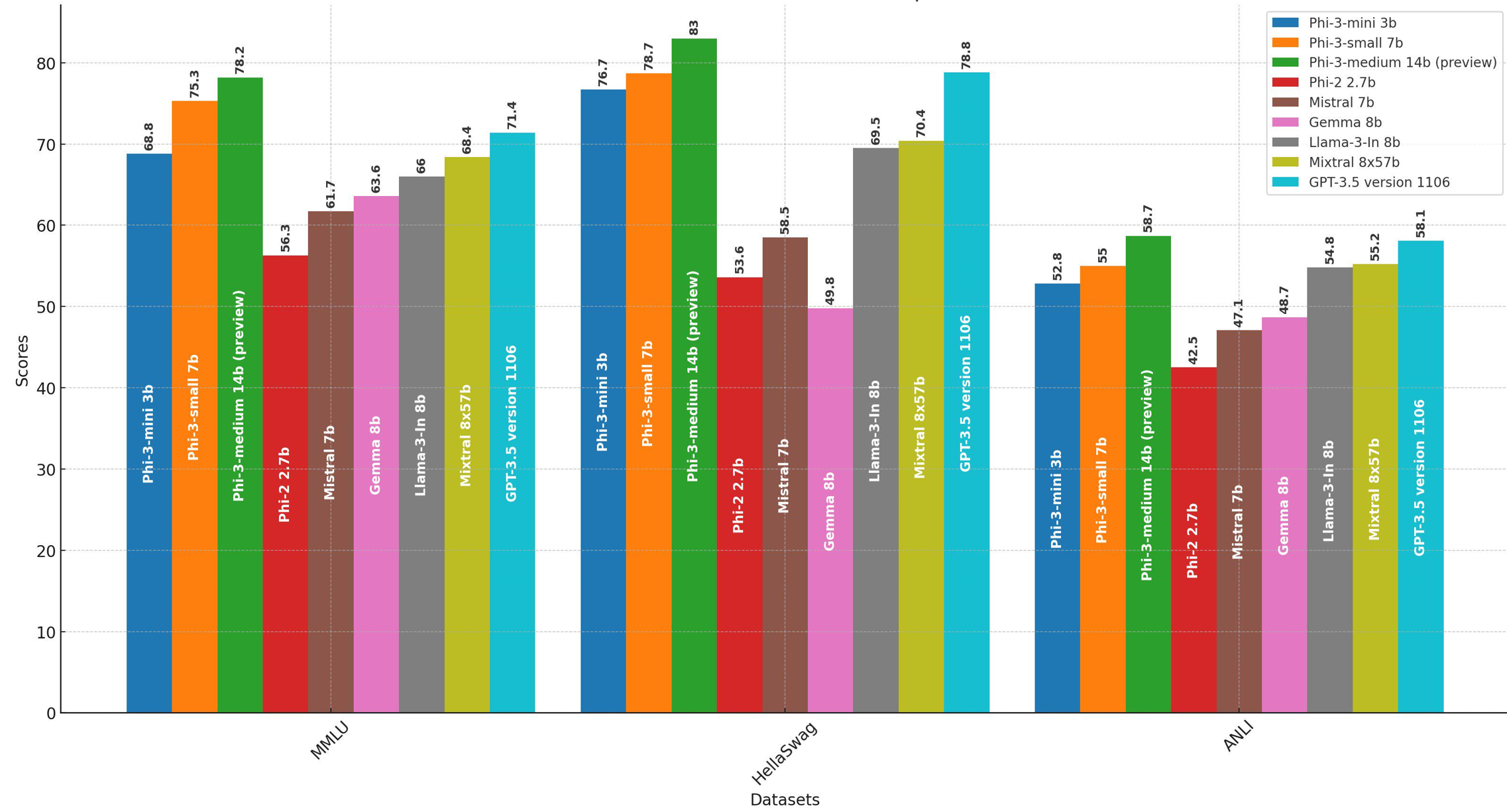} 
  \caption{Comparison of different models.}
  \label{llm_comp}
\end{figure*}

Fig. ~\ref{llm_comp} shows the performances of different models trained with benchmark data. \par
\begin{figure}[h!]
    \centering
    \includegraphics[width=1.0\linewidth]{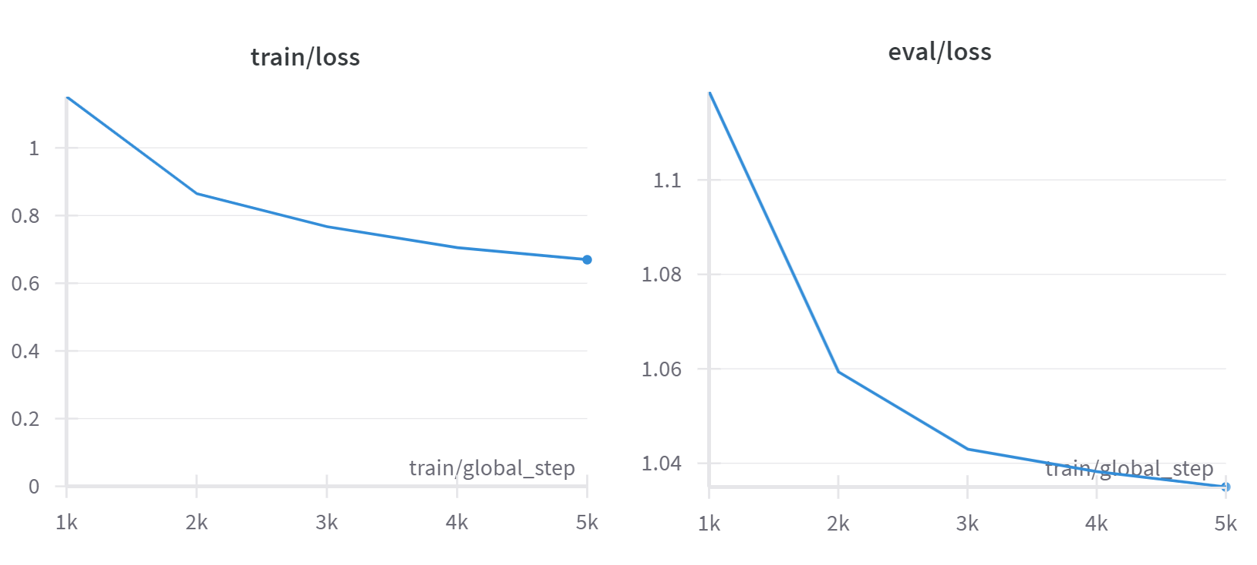}
    \caption{Phi-2 Train and Validation Loss Graphs}
    \label{phi2_loss}
\end{figure}

 As seen from the Fig.~\ref{phi3_loss} and~\ref{phi2_loss} , Phi-3 outperformed the Phi-2 model in all datasets. While the Phi-2 model was trained with 2.7 billion parameters, the Phi-3-Mini model was trained with 7 billion parameters. It can be said that the number of parameters used in the models directly affects the performance shown in the graphs. On the other hand, we know that the number of parameters is directly related to resource usage. Therefore, the Phi-2 model is more appropriate to use in cases where limited resource use is required.\par

In this study, we used the Phi-2-instruct model and the Phi-3-mini model and injected QLoRA into the linear layers of both models. The result we expected was that the answers obtained from the Phi-3 model were better than those obtained from the Phi-2 model. However, the number of trainable parameters in the Phi-3 model is higher than in the Phi-2 model, and when QLoRA is applied to all linear layers, it seems inevitable that the model will underfit or overfit. 
When large models are fine-tuned with small datasets, they may not meet expectations depending on the model structure and the similarity of the fine-tuned data to the original training data. We can say that the same situation was observed for the Phi-3 model in this study. Fig. ~\ref{phi3_loss} shows a gradual decrease in the training loss of the Phi-3, but when we examine the validation loss graph of phi-3, we can see that the model stops learning after a certain step.\par

In Fig.~\ref{phi2_loss}, the training and validation loss graphs of the Phi-2 model are shown. While a decrease in training loss is not as rapid as Phi-3, a gradual decrease is observed. The same decrease can be observed in validation loss. This shows that the Phi-2 model can learn better than the Phi-3 model with a limited dataset. If necessary updates are made to the hyper-parameters used in training, the desired values in the loss graphs can be obtained. If these hyper-parameters; step size, epoch size and batch size, are increased the model can be generalized depending on the training time.\par

When creating the vector database on which the model will be fed in RAG systems, care should be taken to ensure that the data of the platform is consistent and diverse. When creating a vector database, PPT, DOC, JSON etc. Data can be used in various formats, but it is necessary to ensure that this data is transferred to the database correctly.
\begin{table}[b!]
\centering
\caption{Hyper-Parameters and Values}
\label{hyper-parameters}
\renewcommand{\arraystretch}{1} 
\begin{tabular}{|>{\centering\arraybackslash}m{3cm}|>{\centering\arraybackslash}m{2cm}|}
\hline
\textbf{Parameter} & \textbf{Value} \\
\hline
\text{Temperature} & 0.2 \\
\hline
\text{Repetition penalty} & 1.1 \\
\hline
\text{Max new tokens} & 300 \\
\hline
\text{Chunks size} & 200 \\
\hline
\text{Chunks Overlap} & 0 \\
\hline
\end{tabular}
\end{table}

\begin{figure}[h!]
    \centering
    \includegraphics[width=1.0\linewidth]{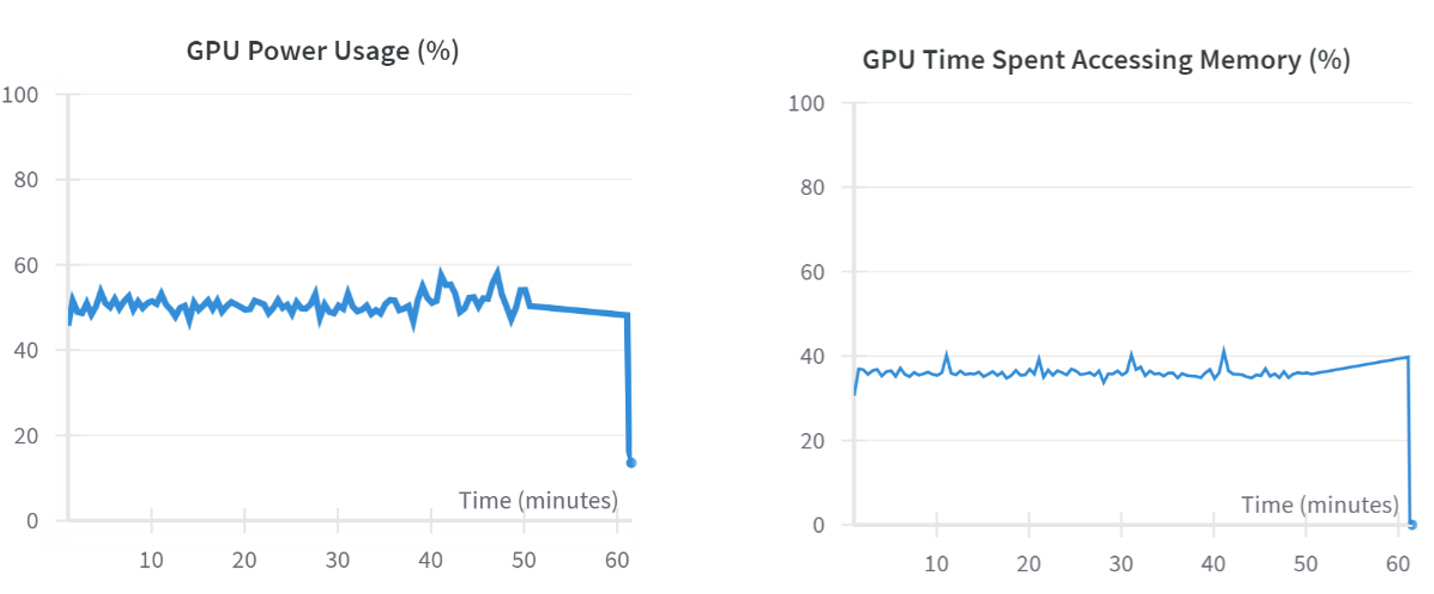}
    \caption{Phi-2 GPU Power Usage}
    \label{gpu_usage}
\end{figure}
Parameters such as chunks size, chunks overlap determined during data transfer and temperature, repetition penalty, and max new tokens used when creating the LLM chain will directly affect content extraction. \par If the temperature is selected high in an application, the randomness in the answers will increase and the answers may move away from reality. The repetition penalty specifies the refresh rate to avoid getting stuck on the same words while extracting information from the context. All of the specified parameters can be optimized and customized according to the targeted system. The hyper-parameters used in this study are given in Table ~\ref{hyper-parameters}.\par

\begin{figure}[h!]
    \centering
    \includegraphics[width=1.0\linewidth]{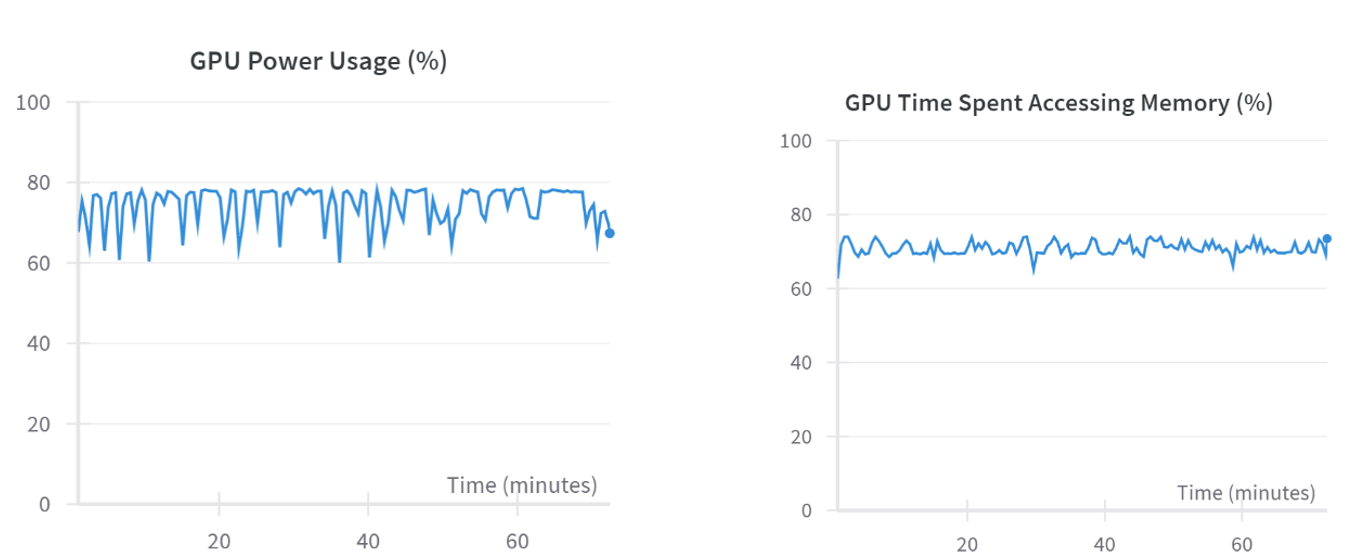}
    \caption{Phi-3 GPU Power Usage}
    \label{phi3_gpu_usage}
\end{figure}

It is known that the GPU requirement of the model will increase depending on the number of parameters. Fig. ~\ref{gpu_usage} and ~\ref{phi3_gpu_usage} shows the GPU Power used by the Phi-2 and Phi-3 models in the same time slot. Although the power used by both models for the 5k step under the same conditions is not stable, the resource required by the Phi-3 model is more than Phi-2. This situation is directly related to the trainable parameters and may vary depending on the number of layers into which QLoRA is injected. In both models used in this study, QLoRA was injected only into linear layers. As can be seen from the graphs, Phi-2 is a more suitable model when working with limited resources.

\section{Discussion \&Conclusion}

In this study, a method is proposed that combines the positive aspects of RAG and SLMs, which are frequently used in the literature. The proposed method is one of the precedents studies on content extraction for vector databases, aspects to be considered when creating prompt content, adding content in different formats to vector databases and observing its effect. \par

In the experiments, in addition to RAG studies, important information was given on many different subjects such as the use of Lora and quantization processes with LLMs/SLMs, the hyper-parameter setting in the fine-tuning process, and the creation of the dataset used in model training. This important information will help researchers on how to use an large language models most effectively for studies that require limited resource use and how to enrich content by reducing storage requirements. \par

Once the Phi-2 and Phi-3 models used in this study were evaluated separately, it was seen that the answers expected from a chatbot could be obtained. However, what is expected from chatbots customized for a domain is to provide clearer and more accurately interpreted answers on specific issues. During the development of the domain-specific chatbot, the RAG unit was added to the system. In the vector database, which was included in the system with RAG, content customized for the talent platform was used to support both SLMs and LLMs. \par

The metrics obtained from the answers given by RAG-supported base Phi models are detailed separately. These metrics represent the accuracy of the answers provided by the model when the base model is used with the vector database. In provided metrics, it is seen that the retrained model outperformed the base models in terms of performance at Unigram, Bigram, Trigram, and 4-gram precision values. The observation that a chatbot performs well in the evaluation based on single and sequential words means it can establish the semantic relationship it extracts from the training set. In this case, it can be said that the trained model can learn with limited data. \par

In the experiments, it is seen that ROUGE values are higher in the RAG-supported chatbot used with the fine-tuned phi-2 model. This value shows how similar the information created by the chatbot is to the reference information in the vector database. In such a case, it is expected that the fine-tuned model will produce results more similar to the vector database. Because in the base model, extra information obtained from the talent platform is not included in the knowledge. This affects the number of words that the chatbot can create, and the number of words has a direct effect on the ROUGE value. The important point to note in the outputs presented in this study is this: ROUGE value close to 1 means getting closer to the reference text, and this may cause overfitting in the retrained models.  In such a scenario, the difference between generalization and overfitting should be well understood when evaluating the trained model. In this study, both the obtained metrics and the training/validation loss graphs show that the model can be generalized without overfitting. \par

One of the important metrics in the evaluation of the domain-specific chatbot proposed in this study is the answers received to the questions asked by the bot. These answers were detailed and evaluated under each experiment. It has been observed that the answers given by each model to different questions and the answers obtained from the talent platform match in terms of content. These question-answer pairs were created by seeing the missing points in existing studies and are examples of similar studies. Diversifying the small/large language models used in the proposed method can contribute to the creation of interactive and reliable chatbots. The outputs obtained in the study are an important continuation of the studies on the more effective and efficient use of chatbots in both academic and industrial applications. 

%






\section*{Data Availability Statements}

The data underlying this article were provided by Huawei under by permission. Data will be shared on request to the corresponding author with permission of Huawei.

\bibliographystyle{IEEEtran}
\bibliography{IEEEabrv,Bibliography}

\pagebreak
\onecolumn

\begin{center}
\section*{\centering Supplementary Material: Efficient Learning Content Retrieval with Knowledge Injection}
\end{center}


\begin{sidewaystable*}[htbp]
    \centering
    \caption{Question \& Answer }
    \small 
    \resizebox{\textwidth}{!}{%
        \begin{tabular}{|>{\raggedright\arraybackslash}p{2.5cm}|>{\raggedright\arraybackslash}p{5.5cm}|>{\raggedright\arraybackslash}p{4cm}|>{\raggedright\arraybackslash}p{4cm}|>{\raggedright\arraybackslash}p{5.5cm}|} 
        \hline
        \textbf{Question} & \textbf{Ground Truth (GT)} & \textbf{Phi-2} & \textbf{Phi-3} & \textbf{RAG+Phi-2} \\
        \hline
        What will I learn in the HCIP course? & The HCIP (Huawei Certified ICT Professional) courses are designed to provide a more in-depth and comprehensive understanding of specific technologies compared to the foundational knowledge covered in HCIA courses. Here’s what you can expect to learn in HCIP courses:
        1. Advanced Routing and Switching
        2. Enhanced Security Measures.
        3. Network Optimization.
        4. Specialized Technologies.
        5. Complex Network Troubleshooting.
        6. System Integration.
        7. Practical Application of Theoretical Knowledge.
        HCIP courses are tailored to prepare professionals for mid-level to senior positions in ICT, enhancing their skills in handling more complex systems and networks. 
     & The HCIP certification exam covers a wide range of topics related to computer hardware, including operating systems, networking technologies, storage devices, and security mechanisms. By passing the exam, individuals demonstrate their proficiency in configuring, maintaining, and troubleshooting enterprise-class computers. This certification validates their skills and expertise in the field, enhancing their career prospects.&
     The HCIP certification course covers key concepts and practical applications of HCIA in the context of enterprise networks. Participants will learn how HCIA addresses routing challenges, its significance in enterprise networks, and its practical implementation using Cisco routers. &In the HCIP course, learners will gain a comprehensive understanding of the history and evolution of computer technology, including key milestones, influential figures, and major advancements. Additionally, students will explore contemporary trends shaping the industry, preparing them for careers in ICT. The course emphasizes critical thinking skills, problem-solving abilities, and practical knowledge applicable to various professional environments. With its interdisciplinary approach, HCIP equips learners with the tools necessary to navigate the ever-changing landscape of information and communication technologies. Whether pursuing further studies or seeking career opportunities, HCIP provides a solid foundation for success in the field. \\
     \hline
     What is the HCIE? & 
     The HCIE, or Huawei Certified ICT Expert, represents the highest level of certification offered by Huawei in its certification program. It is designed for senior professionals in the ICT industry who have extensive knowledge and experience. The certification aims to validate the skills required to construct, optimize, and manage complex network environments. Here's a detailed overview of what the HCIE certification entails:
        Objectives and Focus
        - Expertise: The HCIE certification is aimed at individuals who can demonstrate deep technical expertise in areas like routing and switching, security, wireless, and cloud computing.& 
        HCIE refers to Huawei Certified Expert, a high-level certification indicating advanced expertise in specific Huawei technologies, achieved through rigorous testing and evaluation. HCIE holders possess in-depth knowledge and skills in their chosen area of expertise.& 
        HCIE is the Huawei Certified **Principles and Technologies** exam, which focuses on Huawei's core principles and technologies. It is designed for experienced professionals seeking to deepen their understanding of Huawei's offerings and advance their careers.& 
        The HCIE (HCIP-Cloud Computing) certification validates expertise in designing, implementing, and managing cloud computing solutions. It focuses on practical skills required for cloud infrastructure setup, deployment, and management. HCIE candidates demonstrate proficiency in configuring virtualization environments, deploying cloud services, optimizing resource utilization, ensuring security, and troubleshooting cloud-related issues. This certification prepares professionals for careers in cloud architecture, systems administration, and IT consulting. \\
        \hline
        Can you give a summary of the 'Data Management and Analytics' course?&
        Here's a summary of the "Data Management and Analytics" course:
        Overview: Aimed at students from all majors, this course provides a basic understanding of managing and analyzing data in the digital era using real-life examples.
        Target Audience: Secondary specialized school students, Junior college students, Undergraduates, General audience.
        Recommended Follow-up Course: HCIA-openGauss.
        Course Outline: Data categories, Data management, Data modes, Relational models and operations.
        Course Structure: 9 course modules, 1 experimental module, 1 final exam, 1 course satisfaction survey.&
        The 'Data Management and Analytics' course provides an overview of managing, analyzing, and deriving insights from large volumes of data. It covers techniques and technologies for data storage, retrieval, manipulation, and analysis, as well as methods for visualizing and interpreting analytical results. Participants learn how to apply data management and analytics strategies to solve business problems and make data-driven decisions.&
        The 'Data Management and Analytics' course covers topics such as data modeling, database design, query languages, SQL, data warehousing, data mining, business intelligence tools, data visualization, and big data analytics. Participants learn how to collect, store, process, analyze, and interpret data to support decision-making and improve business processes.&
        The 'Data Management and Analytics' course provides an overview of data management principles, including data modeling, storage, retrieval, and analysis. It covers topics such as relational databases, NoSQL databases, data warehousing, data mining, machine learning, and big data analytics. The course equips learners with practical skills in managing and analyzing data effectively for various applications and industries.\\
        \hline
        \end{tabular}
    }
    \label{question_answers}
\end{sidewaystable*}

\begin{sidewaystable*}[htbp]
    \centering
    \small 
    \begin{tabular}{|>{\raggedright\arraybackslash}p{3cm}|>{\raggedright\arraybackslash}p{5.5cm}|>{\raggedright\arraybackslash}p{4cm}|>{\raggedright\arraybackslash}p{4cm}|>{\raggedright\arraybackslash}p{5.5cm}|} 
    \hline
    \textbf{Question} & \textbf{Ground Truth (GT)} & \textbf{Phi-2} & \textbf{Phi-3} & \textbf{RAG+Phi-2}\\
    \hline
    Why should I pursue the HCIE? &
    Pursuing the HCIE (Huawei Certified ICT Expert) certification can be highly beneficial for several reasons, especially if you are looking to advance your career in the ICT industry. Here’s why you might consider achieving this high-level certification:
    1. Expertise and Skill Enhancement,
    2. Professional Recognition,
    3. Career Advancement,
    4. Personal and Professional Growth,
    5. Networking Opportunities,
    6. Organizational Impact,
    7. Meeting Industry Standards.
    Achieving an HCIE certification is a commitment that involves significant effort and time, but the benefits it brings in terms of career opportunities, professional recognition, and personal satisfaction make it a worthwhile endeavor for ambitious ICT professionals.&
    HCIE certification validates advanced skills and expertise, enhancing career prospects and opening doors to higher-level positions in the field of computer networking. By achieving HCIE status, professionals demonstrate their ability to tackle complex networking challenges and contribute significantly to enterprise infrastructure planning and implementation. &
    The HCIE course in HPC is designed for students passionate about HPC and interested in pursuing careers in HPC system designing, development, and maintenance. It prepares students for HPC certification exams and equips them with practical HPC knowledge and hands-on skills applicable to real-world projects. By completing the HCIE, students can enhance their career prospects in the field of HPC.&
    Pursuing the HCIE can lead to career advancement in the ICT field, as it demonstrates expertise and knowledge in network administration and management. Additionally, HCIE certification enhances employability and opens doors to opportunities in various industries that rely heavily on technology infrastructure. It also provides a solid foundation for further specialization or higher-level certifications in networking and IT management. Overall, HCIE certification is valuable for individuals seeking professional growth and career progression in the ICT industry.\\
    \hline
    How does the '5G Basics: What it's all about' course structure ensure comprehensive learning?&
    The "5G Basics: What it's all about" course ensures comprehensive learning through its structured approach, which includes the following key components:
    1. Modular Structure.
    2. Fundamental Topics.
    3. Practical Applications.
    4. Self-Paced Learning.
    5. Skills Development.
    By combining a comprehensive curriculum with a flexible, self-paced learning format, the course ensures that learners can gain a deep and well-rounded understanding of 5G technology.&
    The 'Introduction to 5G Knowledge' course structure ensures comprehensive learning by covering various aspects of 5G technology, including its history, key features, applications, and impact on industries. Through a combination of theoretical modules and hands-on labs, students gain a holistic understanding of 5G concepts and their practical implications. Additionally, regular quizzes and a final exam assess students' comprehension throughout the course, ensuring they grasp the foundational knowledge required to engage with 5G technologies effectively.&
    The course structure comprising both theory and Hands-on Exercises ensures comprehensive learning. While theory provides foundational knowledge about 5G technology, Hands-on Exercises allow learners to apply this knowledge in practical scenarios, reinforcing understanding and competency.&
    The 'Introduction to 5G Knowledge' course follows a structured approach with modules covering various aspects of 5G technology, including history, architecture, applications, and industry trends. This ensures comprehensive coverage of the topic, providing learners with a well-rounded understanding of 5G. Additionally, the course includes practical exercises and quizzes to reinforce learning and assess comprehension. Overall, the course design aims to equip learners with in-depth knowledge about 5G technology.\\
    \hline
    How do students gain hands-on experience in the 'Principles and Applications of WLAN' course?&
    In the "Principles and Applications of WLAN" course, students gain hands-on experience through several practical approaches designed to enhance their understanding and skills in wireless local area networking. Here’s how this is typically achieved: Lab Sessions, Simulation Tools, Practical Projects, Case Studies, Interactive Demonstrations.&
    Students gain hands-on experience through lab sessions where they practice configuring, securing, and troubleshooting WLANs using network simulators, routers, switches, and wireless devices. These practical exercises reinforce theoretical concepts and develop essential skills for WLAN deployment and management.&
    In the 'Principles and Applications of WLAN' course, students are provided with lab modules involving WLAN design and configuration tasks as well as troubleshooting exercises, allowing them to gain practical experience. Additionally, quizzes are used to check progress and assess understanding throughout the course.&
    Students gain hands-on experience through lab sessions where they can practice configuring, troubleshooting, and optimizing WLAN networks using real-world scenarios. These labs provide opportunities for students to apply theoretical concepts and develop practical skills necessary for WLAN deployment and management.\\
    \hline

    \end{tabular}
    \label{question_answers}
\end{sidewaystable*}

\begin{sidewaystable*}[htbp]
    \centering
    \small 
    \begin{tabular}{|>{\raggedright\arraybackslash}p{3cm}|>{\raggedright\arraybackslash}p{5.5cm}|>{\raggedright\arraybackslash}p{4cm}|>{\raggedright\arraybackslash}p{4cm}|>{\raggedright\arraybackslash}p{5.5cm}|} 
    \hline
    \textbf{Question} & \textbf{Ground Truth (GT)} & \textbf{Phi-2} & \textbf{Phi-3} & \textbf{RAG+Phi-2} \\
    \hline
What types of knowledge or skills do students gain from the '5G Basics: What it's all about', '5G Network and Applications' and 'Introduction to Computer Networks and the Internet' courses?&
5G Basics: What it's all about: Understanding of 5G Fundamentals, Awareness of 5G Features, Introduction to 5G Applications.
5G Network and Applications: In-depth 5G Networking Knowledge, Application-Centric Learning, Network Slicing and Management.
Introduction to Computer Networks and the Internet, Foundational Networking Concepts, Data Communication Techniques, Internet Working.
Combined Skills and Knowledge Outcomes: Technical Proficiency, Problem-Solving Abilities, Innovation and Adaptation.&
Students in the 'Introduction to 5G Knowledge' course gain insights into the history, technologies, standards, and applications of 5G. Those in the '5G Network and Applications' course learn about designing, deploying, and managing 5G networks for various use cases. Meanwhile, participants in the 'Introduction to Computer Networks and the Internet' course acquire foundational understanding of computer networking principles, protocols, and data transmission.&
5G Basics: What it's all about' course provides foundational knowledge about the concepts, technologies, and applications of 5G networks. '5G Network and Applications' delves deeper into the architecture, protocols, and use cases of 5G networks. 'Introduction to Computer Networks and the Internet' covers fundamental topics in computer networking, including routing, switching, protocols, and Internet architecture.&
Students who complete the 'Introduction to 5G Knowledge' course will have a basic understanding of 5G technology, including its principles, applications, and benefits. Participants in the '5G Network and Applications' course will delve deeper into the technical aspects of 5G networks, exploring topics such as network architecture, protocols, and use cases. Additionally, students who complete the 'Introduction to Computer Networks and the Internet' course will gain foundational knowledge about computer networks, including their components, protocols, and functions. This course lays the groundwork for further exploration of networking concepts and technologies.\\
\hline
What career roles can students prepare for after completing the '5G Basics: What it's all about' course?&
After completing the "5G Basics: What it's all about" course, students can prepare for the following career roles:
Telecommunications Engineer, Network Architect, 5G Deployment Specialist, Technology Consultant.
These roles leverage the comprehensive understanding of 5G technology fundamentals, architecture, capabilities, and applications gained from the course.&
After completing the 'Introduction to 5G Knowledge' course, students can prepare for careers in fields related to telecommunications, technology, or infrastructure development. They may pursue opportunities as network engineers, system architects, wireless network specialists, or industry consultants focusing on 5G technologies and applications.&
Real case courses focalizing each major streams leading with real scenarios are embedded into modules through hands-on activities providing inslight on major cloud relevant topics like fundamentalsecularsbasedon real industry demagnrounding.&
After completing the 'Introduction to 5G Knowledge' course, students can prepare for various career roles in the ICT industry, including network engineers, system administrators, software developers, and telecommunications specialists. They will have a solid understanding of 5G technology and its applications, enabling them to contribute effectively to the development and deployment of 5G networks. Additionally, they may pursue further specialization or advanced degrees in ICT to enhance their career prospects.\\
\hline
What is the target audience for the '5G Basics: What it's all about' course?&
The target audience for the "5G Basics: What It's All About" course includes: Telecommunications Professionals, Network Architects and Engineers, Technology Consultants, Students and Academics, IT Professionals, Industry Professionals.&
The 'Introduction to 5G Knowledge' course is designed for students, professionals, and enthusiasts interested in understanding the fundamentals of 5G technology, its applications, and impact on various industries. It caters to individuals with a background in computer science, telecommunications, engineering, or related fields. Additionally, the course is suitable for anyone curious about the future of wireless communication.&
The '5G Basics: What it's all about' course is aimed at students, professionals, and enthusiasts who want to gain a foundational understanding of 5G technology, its significance, and implications.&
The 'Introduction to 5G Knowledge' course is designed for students, professionals, and enthusiasts interested in learning about the fundamentals of 5G technology. It provides a comprehensive overview of 5G principles, architecture, applications, and industry trends, catering to individuals from diverse backgrounds and experience levels. Whether you are studying ICT or not, this course equips learners with essential knowledge to understand and navigate the future of telecommunications. Additionally, it serves as a valuable resource for IT professionals seeking to stay updated with the latest advancements in mobile networking technology. Overall, the 'Introduction to 5G Knowledge' course aims to foster innovation and drive economic growth by empowering participants with the skills necessary to leverage 5G technology effectively.\\
\hline
\end{tabular}
\label{question_answers}
\end{sidewaystable*}
\end{document}